\documentclass[5p, twocolumn]{elsarticle}
\usepackage[utf8]{inputenc}
\usepackage[T1]{fontenc}

\usepackage{graphicx}
\graphicspath{{fig/}}
\usepackage{subfig}
\usepackage{enumitem}
\usepackage{listings}
\usepackage{xcolor}
\usepackage{booktabs}
\usepackage{dirtree}
\usepackage{multirow}

\begin{document}

\title{LAMES: A Large-Scale and Artisanal Mining Environmental Segmentation Dataset}

\author[tum]{Matthias Kahl}
\author[tum]{Zhaiyu Chen}
\author[iitd]{Sudipan Saha}
\author[hanken]{Mrinalini Kochupillai}
\author[orora]{Lukas Kondmann}
\author[tum]{Xiao Xiang Zhu}

\address[tum]{Technical University of Munich, Germany}
\address[hanken]{Hanken School of Economics / Svenska handelshögskolan, Helsinki, Finland}
\address[iitd]{Indian Institute of Technology, Delhi, India}
\address[orora]{OroraTech, Munich, Germany}

\begin{abstract}
 Mining operations are of utmost importance to the economy of some nations. However, such operations result in land-use change, very high energy consumption, and negative impacts on the environment, including soil erosion and deforestation. The mining process can impact an area much larger than the mining site itself. Adding to the negative externalities linked to mining is the fact that, in addition to government-sanctioned legal mining operations, illegal mining is widespread, including in various countries of Africa. The ability to monitor remote mining site activities can be useful, e.g., for the detection of illegal artisanal mining activities and their environmental impacts. An important outcome of such monitoring could include a better understanding of the interrelationship between mine facility attributes (e.g., mining types, processing methods, commodities, etc.) and their impact on the natural environment. In this work, we present a data set that contains 150 Large Scale Mining (LSM) sites and 870\,km² annotated area of Artisanal Small-scale Mining (ASM) sites. The metadata includes nine eminent LSM sections and 27 mining site attributes for each LSM site. We also discuss the data set's possible contribution to the research community, social and environmental consequences, and researchers' responsibilities from an ethics perspective.
\end{abstract}


\maketitle   

  \begin{figure}[t]
    \centering
    \includegraphics[width=\linewidth]{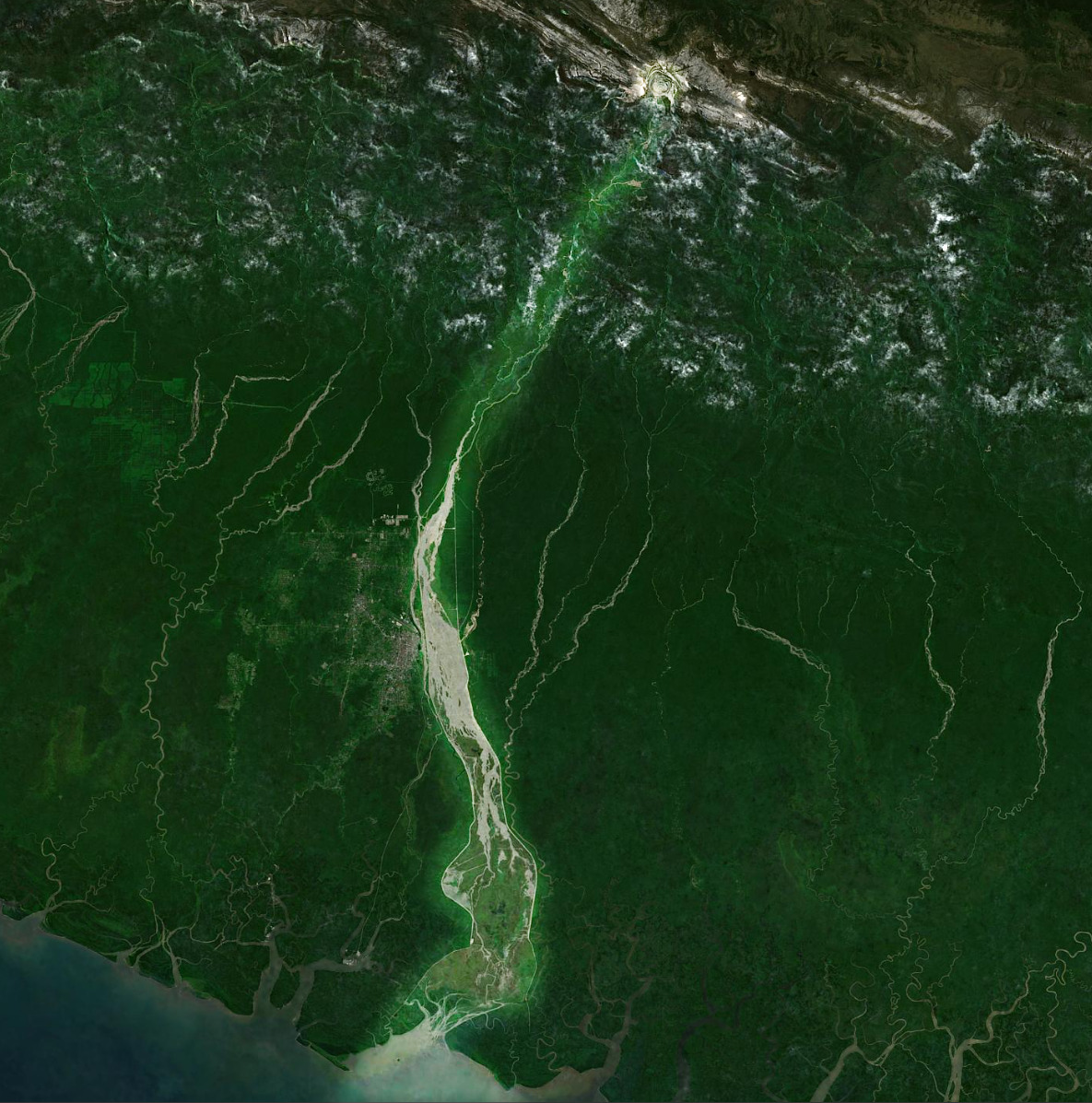}
    \caption{A Sentinel-2 L2 image of the Grasberg mine and its accumulated waste dump in the Indonesian part of Papua. Considering that the image extents roughly equal to 90\,km, one can see the large dimensions.}
    \label{fig:grasbergmine}
  \end{figure}

\section{Introduction}  %% \introduction[modified heading if necessary]
\begin{figure}[t]
    \centering
    \subfloat[Kumasi 29.12.1986 (Landsat 4/5 image)]{
      \includegraphics[width=0.48\linewidth]{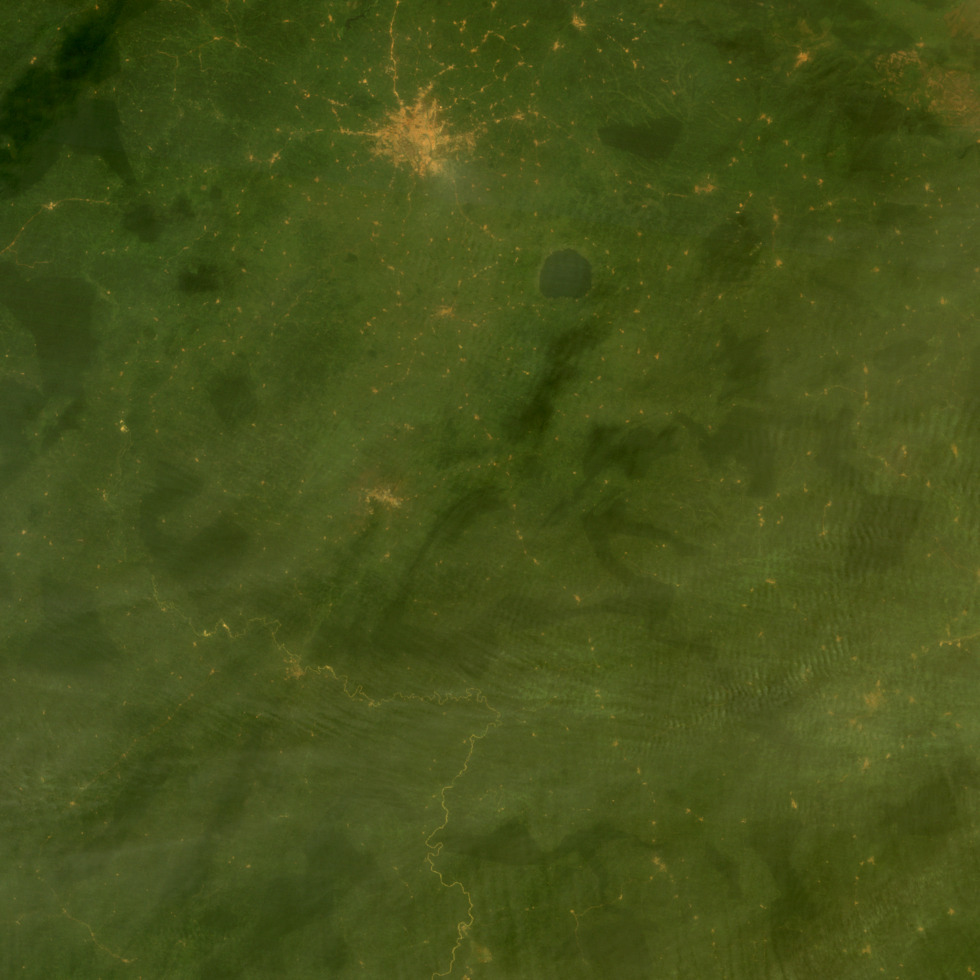}%
      \label{fig:kumasi1986}}%
    \hfill
    \subfloat[Kumasi 22.01.2022 (Landsat 8/9 image)]{%
      \includegraphics[width=0.48\linewidth]{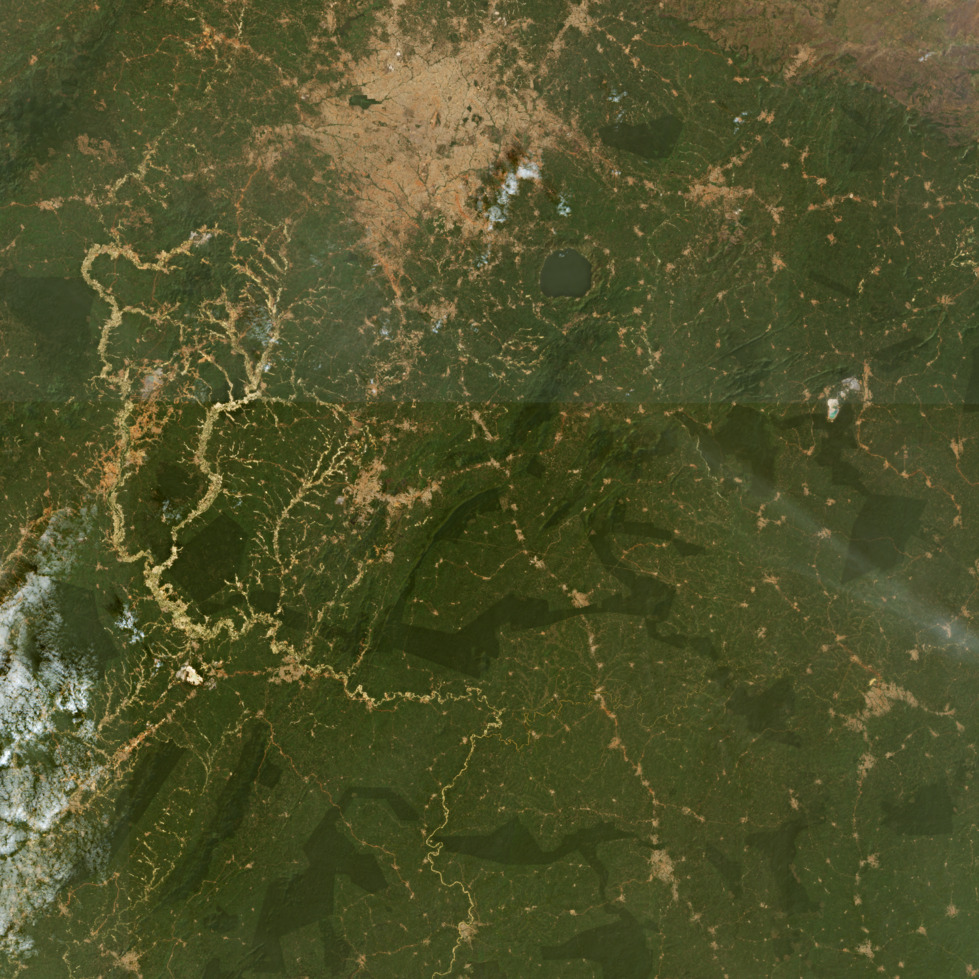}%
      \label{fig:kumasi2022}}%
    \caption{Comparison of Kumasi over time using Landsat images.}
  \end{figure}

  Mining refers to the extraction of valuable minerals and other geological materials from the Earth. In addition to legal mining, illegal mining is also practiced, without legal state permission and in the absence of corresponding rights on mined land in the form of ownership, licenses, or transportation clearances \cite{dozolme2020learn}. Mining operations can have a huge impact on the environment on a local, regional, and global scale. The effects include (I) increased soil erosion \cite{LIU201554} due to deforestation of mining landscapes, (II) ground and surface water pollution with acid or heavy metal  \cite{maest2006predicted} resulting from mine drainage, mine cooling, and deep digging, (III) depletion of biodiversity on land \cite{DIEHL2004} and in water \cite{Price_Tremblay_1993} due to massive environmental changes and tailings, (IV) accumulation of waste from tailings and in the form of spoil tips, and (V) indirect global effects such as increased CO$_2$ concentration in the air from coal mining and burning \cite{BIAN2010215}. A prominent example of the environmental impact can be seen at the Ajkwa river which has been used as a waste dump for the Grasberg mine (see Fig.~\ref{fig:grasbergmine}). The estimated excavated waste of that mine will reach 6 billion tons, or around 1\,km³, or more than twice the amount of moved material on building the Panama canal \cite{riverOfWaste}. The waste already led to 138\,km² loss of forest between 1987 and 2014. The estuarial suspended particulate matter lies above twice the threshold from the Australian guidelines for tropical lowland rivers and estuaries to maintain healthy aquatic communities \cite{alonzo2016capturing}. But also artisanal mining has a large impact on the natural environment. Image~\ref{fig:kumasi1986} and \ref{fig:kumasi2022} show the difference in 35 years around Kumasi in Ghana. The most visible changes are due to the increased population of Kumasi (489,586 in 1984 vs. 2,907,000 in 2017) \cite{ghanastat} and the growth of Artisanal Small-scale Mining (ASM) sites in the area south of the City.

Thus, it is important to regulate and restrict mining activities to minimize negative externalities. Yet, due to economic expediencies and complex political and human rights issues, often, governmental authorities tolerate, turn a blind eye towards, are not aware of, or fail/refuse to effectively sanction illegal mining activities and their environmental impacts. One way to effectively avoid this inaction is by gaining public awareness of those activities. Public awareness should be based on publicly available information sources to avoid manipulated or false information. Mining activities need to be detected and classified as legal or illegal through transparent authorization steps or processes. Monitoring mining activities including mine size, excavated volume, deforestation, and waste volume allows detection of larger violations of permitted thresholds in an early stage.

Timely and accessible monitoring of mining activities is now possible thanks to the increasing commercialization and widespread availability of global remote sensing technologies, such as the Copernicus program (e.g., Sentinel-1, Sentinel-2, Sentinel-5p). Their frequent revisit of min. every 5 days \cite{spoto2012overview} and their multi-spectral sensors allow tracking of changes in a sufficient temporal and spatial resolution. We would like to encourage researchers to investigate the possibilities of AI-based mining monitoring by contributing with:

\begin{itemize} \setlength\itemsep{0mm}
    \item geo-spatial Chilean LSM sector data set
    \item geo-spatial Ghanaian ASM sites data set
    \item prepared masks for segmentation tasks
    \item mining sector classification experiment
    \item mining site detection experiment
\end{itemize}

%The purpose of this work is to identify mining facilities, mineral classes, and approximation of the mining scale and their environmental impact, based on publicly available Remote Sensing images.
In section \ref{background}, we study the applications of remote sensing data for exploring minerals and the impact of mining on the natural environment as well as on mining artisans' livelihoods and show existing mining monitoring approaches as well as remote sensing suitable mining data sets. In section \ref{datasetObjectives}, we briefly discuss the requirements and targets of this work, while practical data set information and possible applications are provided in section \ref{datasetRecords} and \ref{useCases}.

%%%%%%%%%%%%%%%%%%%%%%%%%%%%%%%%%%%%%%%%%%%%%%%%%%%%%%
\section{Background \& Related Work}\label{background}

\subsection{Mineral exploration}\label{mineralExploration}

Effective mineral deposit mapping is important for sustainable mining. As discussed by Chandrasekar et\,al. \cite{chandrasekar2011investigation}, in absence of proper mineral deposit mapping, beach sands from the southern coast of India are exported entirely in raw form, thus harming the environment.

Optical multispectral images are generally used in the survey stage to rapidly detect promising mineralized zones at a low cost. Towards this, both Sentinel-2 and Landsat sensors are useful \cite{adiri2020recent,sabins1999remote}. In one of their first works, Sabins \cite{sabins1999remote} used Landsat thematic mapper (TM) ratio images for mineral exploration. Chandrasekar et.\,al. \cite{chandrasekar2011investigation} performed a study based on Landsat-7 ETM data to perform mineral mapping with ENVI's spectral angle mapper (SAM). This method determines the similarity between two spectra by calculating the spectral angle between them. Zoheir and Emam \cite{zoheir2012integrating} exploited several simple techniques like band rationing, principal component analysis (PCA), false-color composition (FCC), and frequency filtering (FFT-RWT) of ASTER and ETM+ data for visual interpretation for detailed mapping of the Gebel Egat area in South Eastern Desert of Egypt. Their work further exploits the USGS spectral library (https://speclab.cr.usgs.gov/) of rock-forming minerals for mineral exploration. Mielke \textit{et. al.} \cite{mielke2014potential} investigated potential applications of Sentinel-2 in mineral exploration. 
Ge \textit{et. al.} \cite{ge2020assessment} assessed the capability of Sentinel-2 imagery for iron-bearing mineral mapping. Hyperspectral images are generally used to map minerals in more detail. Advanced Spaceborne Thermal Emission and Reflection Radiometer (ASTER) is used in \cite{bishop2011hyperspectral} for hydrothermal mineral alteration exploration. Zadeh \textit{et. al.} \cite{zadeh2014sub} studied sub-pixel mineral mapping of a porphyry copper belt using EO-1 Hyperion data. A comprehensive review of ASTER and Hyperion sensors for mineral exploration is provided in \cite{pour2014aster}. In addition to optical sensors, active Synthetic Aperture Radar (SAR) sensors have also been used for mineral exploration, e.g., for gold mapping \cite{pour2014exploration}.

\subsection{Impact on Natural Environments} \label{impact}

In this section, applications of multi-temporal remote sensing data for analyzing mining impacts are discussed:

\textit{Vegetation stress}: Vegetation is an important part of the environment and also an indicator of ecological integrity. Mining activities severely impact vegetation. Vegetation stress can be monitored 
by monitoring popular vegetation indices, e.g., NDVI, RedEdgeNDVI \cite{makinde2015spectral, xia2007vegetation}. \textit{Analyzing impact on water:} Mining severely impacts water bodies due to discharges of mining tailings into water bodies present near the mining site. Lobo et.\,al. \cite{lobo2015time} proposed an approach to estimate the Total Suspended Solids (TSS) of the water of the Tapajos River using Landsat image time-series. This study spans a time interval from 1973-2013 and uses different kinds of Landsat sensors. Their analysis is based on the assumption that a spectrally invariant target is available near the waterbody that can be used as a reference target to optimize atmospheric correction. In their work, Amazonian dark dense forest spectra present near the Tapajos river basin were used as such an invariant reference. Based on their analysis, they proposed a regression between TSS and surface reflectance. Schmidt and Glasser \cite{schmidt1998multitemporal} demonstrated identification of different water classes within the surface mine area that they postulated to be caused by different pH values, Fe2+ and Fe3+ concentrations, and differences in suspended material.

\textit{Acid mine drainage}: Acid mine drainage occurs when metallic ores from mining activities are exposed to air and water. This causes severe pollution of ground and surface water (and thus is related to discussions in subsection ``analyzing the impact on water''). \textit{Fire caused by mining:} Surface and subsurface fire in the mining sites (especially coal) is an important problem related to mining. Mishra et\,al. \cite{mishra2011detection} proposed an approach for coal mine fire detection in Jharia coalfield (India) using Landsat-7 ETM+ sensor. They exploited the band 6: thermal (10.40 - 12.50 µm) band of ETM+ sensor to estimate surface temperature that is used as an indicator of fire. Huo et\,al.. \cite{huo2014detection} proposed an approach using Landsat ETM+ sensor to study fires in Rujigou coalfield, China. They found ETM+ band 7: thermal and band 6 (SWIR) to be useful for fire detection in coal mining areas. \

\textit{Quantifying surface mining activity and reclamation:} It is important to monitor the expansion of mining activities and reclamation of mining lands. Petropoulos et al. \cite{petropoulos2013change} proposed a supervised multi-temporal approach to monitor them using Landsat images. Their approach is based on a support vector machine classification into the following classes: vegetation, bare soil, water, and active mines. In the next step, change detection is performed based on the post-classification approach. 

\textit{Illegal mining and mining conflict detection}: Some mineral-rich countries, especially those in Africa, are prone to illegal mining activities and conflicts related to mining. While illegal mining activity detection is somewhat related to quantifying surface mining activity, detection of illegal mining needs to work under the assumption that very little / no ground truth data is available \cite{schoepfer2010monitoring, luethje2014geographic}. Luethje et.\,al. \cite{luethje2014geographic} studied one such area in the Democratic Republic of Congo and proposed a multi-scale multi-temporal approach based on the Geographic Object-Based Image Analysis (GEOBIA) technique for detecting illegal mining activities.  In the first step, they used hi-res images (RapidEye, 6.5 meter/pixel resolution) to identify potential mining areas. Based on the identified potential areas (region of interest), a more detailed multi-temporal analysis revealing changes over time was conducted based on very hi-res (GeoEye-1: 0.5 meter/pixel and Ikonos: 1 meter/pixel) images. Post-classification was the change detection approach used in the study.

\textit{Land subsidence and tailing dams stability}: Land subsidence is a common phenomenon around mining areas, which causes terrain instability and endangers the ground construction facilities. However, land subsidence and tailing dam stability monitoring are generally based on SAR-based interferometry, which is out of the scope of this literature survey \cite{yuan2017mining}. Interferometry-based analysis can be supported by multi-temporal multi-spectral visualization (however, this needs further investigation).

\subsection{Impact on Artisans Livelihood}\label{artisanslivelihood}

Livelihood is defined as the way someone earns the money needed to pay for the basic amenities of life, such as food, shelter, and clothing (Cambridge Dictionary). Under human rights law, the right to livelihood, also referred to as the right against deprivation of one’s means of subsistence, is closely associated with the right to life and the right to land, including the right against illegal evictions from (ancestral) land.

The ethical issues linked to earth observation based monitoring of ASM, and its impact on artisanal livelihood have been extensively discussed earlier by \cite{9954451}. We summarize the key points here. A broader ethical perspective in the context of geoscience and the usage of AI is discussed in \cite{cleverley2024ethical}.

Globally, an estimated 40 million people are employed in artisanal and small-scale mining activities. Currently, ASM activities are conducted by four broad categories of people and communities: 

\begin{itemize} \setlength\itemsep{0mm}
    \item Traditional usage of land for ASM (MacDonald 2006);
    \item Former or current small-scale farmers, to enhance their income (Okoh and Hilson, 2011);
    \item Displaced LSM workers (Baah-Ennumh and Forson, 2015)
    \item Government-authorized companies. 
\end{itemize}

In a decided court case, the African Commission on Human and People's Lives saw the traditional use of land as a more important consideration to bear in mind than the status of being “indigenous” or not. According to the court decision, communities that have traditionally used the land for ASM can legally continue using land for this purpose, unless alternative sources of livelihood that also secure their traditional and cultural ways of life are offered and accepted. The court also said that: "The fact that the victims cannot derive their livelihood from what they possessed for generations means they have been deprived of the use of their property under conditions which are not permitted under Article 14.” \cite{chenwi2010african} This means that the remaining three categories of ASM activities are also protected by the right to livelihood if miners have been conducting such activities for "generations".

Notwithstanding the above human rights-based arguments that speak against the forced termination of unauthorized ASM, the above-described negative (environmental) externalities linked to ASM bring us to a kind of "ethical dilemma" where policymakers need to choose between securing the livelihood of those traditionally engaged with ASM for a living and securing environmental health and the health of ASM workers themselves. In this context, research also reveals that ASM, while contributing significantly to the national economy of various countries, including Ghana, keeps artisans caught up in a “poverty trap” \cite{ijerph120708133}.

Permitting or not permitting ASM, therefore, is akin to an ethical dilemma that has no immediate solution. Because ASM makes a significant contribution to the GDP of the countries where it is widely practiced, the governments face a conflict of interest when pressurized to curb or punish illegal mining. In this scenario, there is an urgent need for governments to create alternative economic and development opportunities for artisans currently caught in a poverty trap with ASM being their only source of subsistence. This is also an urgent mandate closely linked to the UN Sustainable Development Goal “Zero Poverty”. 

\subsection{Mining Monitoring}\label{miningMonitoring}
A few works have exploited Copernicus images for mining monitoring \cite{lobo2018mapping, yuan2017mining}. Lobo et.\,al. \cite{lobo2018mapping} proposed a supervised method based on Classification and Regression Trees (CART) applied on Sentinel-2 images for mapping mining areas in the Brazilian Amazon. Yuan et\,al. \cite{yuan2017mining} proposed a method for land subsidence monitoring using Sentinel-1 data. An open data alternative before Copernicus is Landsat data which has, for example, been employed to monitor mining wastewater \cite{rudorff2018remote}. Beyond medium-resolution data, several approaches exist that make use of high-resolution data for monitoring mining activities \cite{chen2020fine,isidro2017applicability}. A variety of contributions combine several data sources and map mining activity or mining potential from several sensors that often vary in resolution \cite{sekandari2020application,zoheir2019multispectral,zoheir2019orogenic}. Further, monitoring mining activities with remote sensing data is the goal of several commercial or humanitarian projects. The MAA project\footnote{https://maaproject.org/2021/mining\_rivers/} analyses gold mining activities in Peru to determine a potential association with deforestation. Similarly, ASMSpotter\footnote{https://dida.do/asmspotter-demo} segments potentially illegal gold mines in Peru, Suriname, and Guyana based on Sentinel-2 data.

The increasing availability of Copernicus data may be particularly advantageous for mining monitoring research in the future for several reasons:

\begin{enumerate}
  \item Increased range of products: Sentinel 5p can be useful for mining-related air pollution monitoring. Sentinel-1 can be used for land subsidence monitoring and Sentinel-2 multi-spectral bands can be useful for mining activity detection and multi-temporal analysis.
  \item Improved data quality due to novel sensor technology, e.g. Sentinel-2 poccesses improved spectral, spatial, and temporal resolution compared to Landsat missions.
  %density of spectral, spatial, and temporal information for Sentinel 2 compared to Landsat missions. 
  \item Its open and free data policy democratized the applicability of satellite-based Earth observation for practical problems.
  %Benefit from community-based approaches due to publicly available images and data.
  \item Long-term perspective: The Copernicus program ensure reliable data access till 2030 and further continuation plan is under development.
\end{enumerate}

These opportunities in mining research have, however, not been fully exploited due to a lack of publicly available datasets.

\begin{figure}[htbp]
  \centering
  \includegraphics[width=\linewidth]{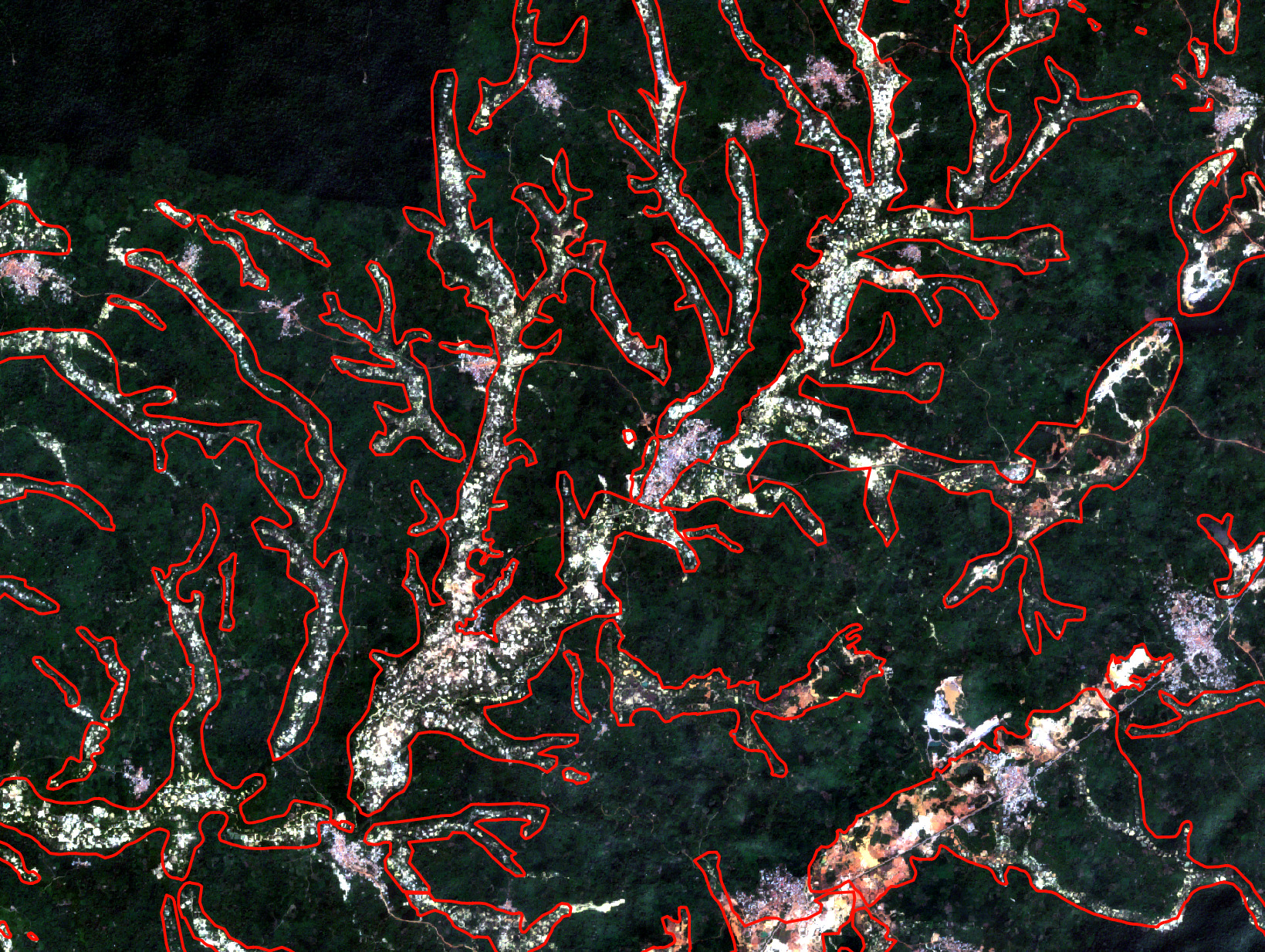}%
  \caption{Typical form of ASM sites along flowing water bodies.}
  \label{fig:asmform}
\end{figure}

\subsection{Mining data sets}
  
  \label{miningDataSets}
  Despite several satellite-image-based studies on mining, most of them use privately-owned data/annotations. This significantly hinders research work in this direction. Despite this, a couple of public data sets exist. One such data set has been introduced in \cite{maus2020gmpv}, by delineating mining areas within a 10\,km buffer from the geographical coordinates of more than 6,000 mining sites across the globe. Their data set consists of 21,060 polygons adding up to 57,277\,km².
  
  Balaniuk \cite{balaniuk2020mining} exploited deep learning for mining and tailings dam detection. They used locations of officially registered mines and dams from the Brazilian government's open data resource. Such data is often available from other governments, too. As an example, a  national-scale geospatial database showing the most important mines, mineral deposits, and mineral districts of the United States is provided.\footnote{https://mrdata.usgs.gov/} The geological distribution of all onshore mineral resources in countries including the United Kingdom is available for purchase.\footnote{https://www.bgs.ac.uk/datasets/bgs-mineral-resources/} Even though such resources are available, they are not provided in a format that is ready for use by data scientists.

%%%%%%%%%%%%%%%%%%%%%%%%%%%%%%%%%%%%%%%%%%%%%%%%%%%%%%
\section{Sources and Processing}\label{datasetRecords}

  Herewith, we provide a data set that facilitates LSM and ASM monitoring with remote sensing technologies by combining LSM metadata and hi-res annotations. The data set comprises eminent sections of LSM facilities located in Chile, as well as the majority of ASM site structures in Ghana. 
  
  The data set supports researchers' understanding of the mining life cycles and impact on the natural environment. To provide high annotation precision and a substantial degree of mining site details for each LSM facility, high knowledge in the field of mining and mineral exploration is necessary. As a consequence, we consulted the \emph{Institute of Mineral Resources Engineering} of the RWTH Aachen as experts in the research field of mining technologies for the annotation process.

  \subsection{Data Sources and Formats}
  To create the dataset, we used the following data sources:

  \textit{Mining Site locations:} In order to select representative LSM sites, we analyzed the Global Mining Polygons data set from \cite{maus2020gmpv} and defined the mining site locations as annotation starting points.

  \textit{Satellite Imagery:} The annotation process is based on hi-res true color satellite imagery from Maxar and CNES. The imagery is provided by Google Earth. The mining sector classification experiment is applied to the hi-res imagery, while the mining site detection experiment is applied to the mid-res imagery from Copernicus Sentinel-2.

  \textit{Country shape:} The country shape for Chile has been retrieved from the natural earth admin-0 vector map \cite{naturalearth_admin0_2024}

  \textit{Mining Site Metadata:} The main sources for this information is MDO\footnote{https://miningdataonline.com/}, Mining News\footnote{https://www.miningnewsfeed.com/}, Diggings\footnote{https://thediggings.com/}, Antofagasta Minerals\footnote{https://www.aminerals.cl/} and Centro Nacional de Pilotaje\footnote{https://pilotaje.cl/}

  %OK - 150 mining sites 
  %OK - sectors
  %OK - country
  %OK - google satellite images / CNES and Maxar
  %OK - Google Earth
  %- kml -> converted to geojson
  %- mining site metadata sources (appendix)

  % \subsection{LSM Metadata and ASM search region}\label{metadata}  

\begin{figure}[htbp]
  \centering
  \includegraphics[width=\linewidth]{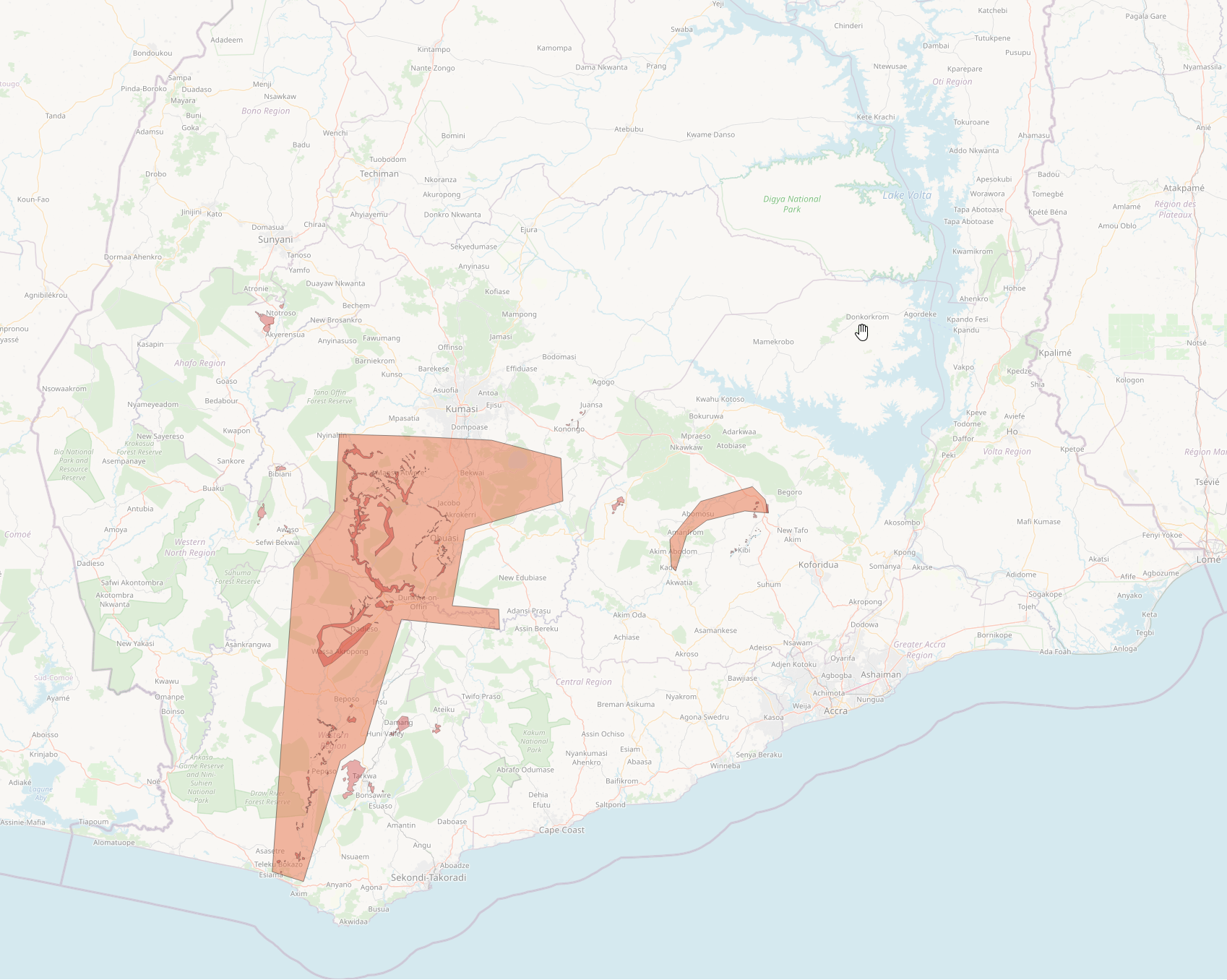}%
  \caption{Image of the ASM site density (red) and the finally defined ASM search regions (orange) in Ghana.}
  \label{fig:asmsearcharea}
\end{figure}

\subsection{Mining Site Selection}\label{dsLSM}

  The search region for ASM sites is defined on visual inspection, the density of clearly visible ASM structures, and previously annotated mining polygons of the Global-scale mining polygons data set from Maus~et.\,al.\cite{maus2020gmpv}. The final extent of the search area can be seen in Fig.~\ref{fig:asmsearcharea}.
  
  Regarding LSM, the focus lies on a representative set of mining sites in Chile. Chile has a quite prominent mining industry across a large range of climate regions with most mining sites in mostly dry regions, making it a suitable country for the dataset due to high availability of cloud-free imagery for the dataset and potential follow-up work. The rough extents of the most known mines in Chile are retrieved from the global-scale mining polygons dataset of \cite{maus2020gmpv}. As of writing this paper, the dataset depicts 228 areas/polygons related to mining in Chile. To keep the annotation effort of <\,1,000\,km² on a reasonable scale, still keeping a representative number of mining sites (150), any mine in Chile larger than 80\,km² has not been considered for the annotation process. Since several mining sites have an overlap that would interfere with segmentation experiments, we combined the 150 mining sites to 71 non-overlapping sites.

  An area of 963\,km² distributed to 150 LSM mining sites is being annotated. The data set from \cite{maus2020gmpv} includes 227 individual mining sites or areas with a mean size of 16\,km² and a standard deviation of 105. As the numbers allow for interpretation, the range of mine facility extents is very broad. The five largest mines cover already 70\,\% of the area of all mines in Chile in the data set of \cite{maus2020gmpv}. The largest mine comprises 1,354 km². We decided, based on the individual provided mine extents, to select the mines in a compromise between including large mines, small mines, and a representative number of mines. Therefore, we dropped the 8 largest and 77 smallest mines, allowing us to annotate the sectors of 150 mining sites, with a total area of 963\,km², mostly in the regions Tarapaca, Antofagasta, Atacama, Coquimbo, Libertador General Bernardo O'Higgins, and Metropolitana.

\subsection{Ground Truth Annotations}\label{annotations}
  The labels of the LSM sectors and ASM areas as polygons represent the heart of our data set. The annotation process includes the key sections of prominent mining sectors within our considered LSM sites in Chile, followed by a detailed metadata investigation for each of these sites. Additionally, it includes the examination and refinement of artisanal and small-scale mining (ASM) annotations based on the work of Maus et al. (2020), with updates and redrawing performed to a higher level of detail. Lastly, the task also entails identifying and annotating ASM sites in Ghana that were not covered by Maus et al. in their previous study.
  
  The annotations are provided in three different \texttt{*.geojson} files. The files are projected in \texttt{EPSG:4326}.

  \begin{itemize}
    \item \texttt{LSM\_Chile\_extents.geojson} Each LSM-site of this work corresponds to a GeoJSON feature with the original polygon extents and IDs from the data-set of \cite{maus2020gmpv}.
    
    \item \texttt{LSM\_Chile\_sectors.geojson} Each of the 1,204 LSM-site sectors of this work corresponds to a GeoJSON Feature. In addition to the polygon sector shapes, the corresponding mining site ID as used in the data set of \cite{maus2020gmpv} is provided. Furthermore, the sector class ID according to Table~\ref{tab:classes}, as well as the acquisition date and acquisition agency are provided.
    
    \item \texttt{ASM\_Ghana.geojson} Since ASM sites grow along flowing water-bodies and have no defined borders or extents, the GeoJSON \texttt{features} encapsulates connected mining site areas from non-connected ones, rather than any enclosed mining site.
  \end{itemize}

  \subsection{Preparation for Semantic Segmentation Tasks}\label{preprocessing}
  The specific task of semantic segmentation in the field of computer vision usually requires original and naturally appearing images that contain the desired segments or objects and segmentation masks that define the class membership on a pixel level. Therefore, for each considered sector, the image set consists of:

  \newcommand{\imgHeight}{40mm}
  \newcommand{\imgWidth}{0.46\linewidth}
  
  \begin{figure}    
    \centering
    \subfloat[Sentinel-2 TC\\GSD:~10\,m]{\includegraphics[width=\imgWidth]{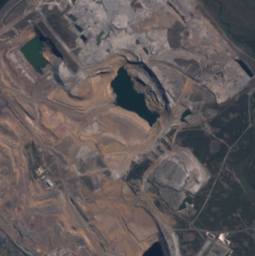}}%
    \hfill
    \subfloat[Binary Mask\\GSD:~10\,m]{\includegraphics[width=\imgWidth]{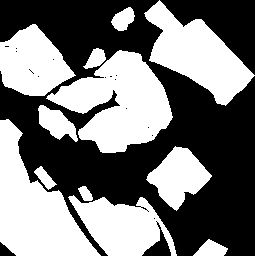}}%
    \hfill
    \subfloat[HiRes TC\\GSD:~0.5\,m]{\includegraphics[width=\imgWidth]{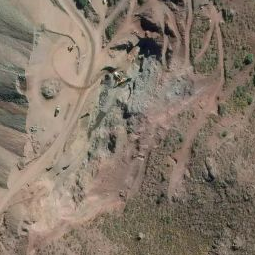}}%
    \hfill
    \subfloat[HiRes Mask\\GSD:~0.5\,m]{\includegraphics[width=\imgWidth]{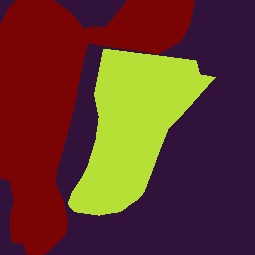}}%
    \caption{The figure shows sample imagery from the mining site detection data set (left) and the mining sector classification data set (right).}%
  \end{figure}

    %\begin{itemize} \setlength\itemsep{0mm}
    %    \item 13 gray-scale segment bounding box images
    %    \item 13 gray-scale segment only images
    %    \item 1 true color segment bounding box image
    %    \item 1 true color segment only image
    %    \item 3 segmentation mask images (10\,m, 20\,m, 60\,m)
    %\end{itemize}
    
  Depending on the implemented algorithm, a dimensional homogenization with padding and/or cropping of the images and masks might still be necessary, since each segment is naturally of a different size.

%%%%%%%%%%%%%%%%%%%%%%%%%%%%%%%%%%%%%%%%%%%%%%%%%%%%%
\section{Data Set Structure}\label{datasetObjectives}  

  \subsection{Scope of supply}
   The data set includes the following content:

      \begin{itemize}
       \item Polygon annotations for LSM (Chile) and ASM in (Ghana)
       \item LSM metadata and sources
       \item Training, Validation and Test Site Polygons
       \item Annotation Documentation
       \item Sentinel-2 patches for Mining Site Detection
       \item QGIS project file for project visualization
       \item Result report of mining site detection and sector classification
   \end{itemize}

    \begin{table*}[htbp]
        \caption{The folder structure and its contents.}
        \setlength{\tabcolsep}{7pt}
        \renewcommand{\arraystretch}{1}
        \footnotesize
        \begin{tabular}{llll}
        \toprule
        Content                            & What                       & Format                    & Count                           \\ \midrule
        \footnotesize \texttt{annotations/Chile\_LSM\_sectors.geojson}  & LSM Annotation            & GeoJSON        & 1,204 Polygons                  \\
        \footnotesize \texttt{annotations/Chile\dots Maus\_et\_al\_subset.geojson}                  & LSM Footprints & GeoJSON        & 227 Polygons   \\
        \footnotesize \texttt{annotations/Ghana\_ASM.geojson}     & ASM Annotation                  & GeoJSON        & 1,288 Polygons                  \\
        \footnotesize \texttt{annotations/overview.qgz}           & Map Visualization               & QGZ            & 6 Layer                         \\
        \footnotesize \texttt{annotations/train\_sites.geojson}   & Train Sites (site detection)    & GeoJson        & 104 Polygons                    \\
        \footnotesize \texttt{annotations/test\_sites.geojson}    & Test Sites (site detection)     & GeoJson        & 35 Polygons                     \\
        \footnotesize \texttt{annotations/doc.pdf}                & Annotation \& Sector            & PDF            & 4 pages                         \\
        \footnotesize \texttt{img\_sector/*}                      & Patches \& Masks (GSD: ~.5\,m)  & PNG            & each 253,731 patches            \\
        \footnotesize \texttt{img\_site/*}                        & Patches \& Masks (GSD: 10\,m)   & GeoTiff        & each 495 patches                \\
        \footnotesize \texttt{metadata/mine\_sites.xlsx\,|.csv}   & Table and Figures               & XLSX, CSV      & 150 entries, 29 Metrics         \\
        \footnotesize \texttt{metadata/sources.txt}               & Webpages                        & Links          & 37 entries                      \\
        \footnotesize \texttt{example Images/*}                   & Images \& Masks                 & PNG, GeoTiff   & 4 PNG, 2 GeoTiff                \\
        \footnotesize \texttt{results/*}                          & Tables and Metrics              & PNG, GeoTiff   & 4 PNG, 2 GeoTiff                \\
        \bottomrule
        \end{tabular}
    \end{table*}

    \begin{table}[t]
    \centering
    \footnotesize
    \caption{Used Image data in \texttt{./site\_detection/}}
    \setlength{\tabcolsep}{2.5mm}
    \renewcommand{\arraystretch}{1.2}
    \begin{tabular}{llrr}
      \toprule
       Set                   & ~~~Path                                & \# Files & Resolution  \\ \midrule
      \multirow{2}{*}{Train} & \texttt{./patches\_img\_trainset/}     & 168,510  & 256x256x3   \\ 
                             & \texttt{./patches\_mask\_trainset/}    & 168,510  & 256x256x1   \\
                             \midrule
      \multirow{2}{*}{Valid} & \texttt{./patches\_img\_validset/}     & 41,591   & 256x256x3   \\
                             & \texttt{./patches\_img\_validset/}     & 41,591   & 256x256x1   \\
                             \midrule
      \multirow{2}{*}{Test}  & \texttt{./patches\_img\_testset/}      & 43,630   & 256x256x3   \\
                             & \texttt{./patches\_img\_testset/}      & 43,630   & 256x256x1   \\
      \bottomrule
    \end{tabular}
    \label{tab:site_image_data_sets}
  \end{table}  

  \begin{table}[htbp]
    \centering
    \footnotesize
    \caption{Image data in \texttt{./sector\_classification/}}
    \setlength{\tabcolsep}{1.5mm}
    \renewcommand{\arraystretch}{1.2}
    \begin{tabular}{llrr}
      \toprule
       Set                   & ~~~Path                                & \# Files & Resolution  \\ \midrule
      \multirow{4}{*}{Train} & \texttt{./mines\_train/}               & 109      & various     \\ 
                             & \texttt{./mines\_train/patches/}       & 361      & 256x256x3   \\
                             & \texttt{./mines\_train/masks/}         & 109      & various     \\
                             & \texttt{./mines\_train/masks/patches/} & 361      & 256x256x1   \\ \midrule
      \multirow{4}{*}{Test}  & \texttt{./mines\_test/}                & 40       & various     \\
                             & \texttt{./mines\_test/patches/}        & 134      & 256x256x3   \\
                             & \texttt{./mines\_test/masks/}          & 40       & various     \\
                             & \texttt{./mines\_test/masks/patches/}  & 134      & 256x256x1   \\ \midrule
      \multirow{2}{*}{Chile} & \texttt{./CHILE/}                      & 1,842    & 2304x2304x3 \\
                             & \texttt{./CHILE/patches/}              & 148,881  & 256x256x3   \\ 
      \bottomrule
    \end{tabular}
    \label{tab:sector_image_data_sets}
  \end{table}

  \subsection{Formats}
  The commercial information of the LSM sites is provided in the common \texttt{.xlsx} spreadsheet format as well as a semicolon-separated values \texttt{*.csv} file for automated reading.

  Besides that, we included the training and test-site footprints, the nearest site neighbor distances, a dockerfile for environment creation, the footprints of all sentinel tiles as well as source files for training, testing, prediction, optional augmentation, creation of footprints, Sentinel 2 image downloading via WCS, image patch extraction \& merging, and random sampling. Table~\ref{tab:site_image_data_sets} shows the data sets image subsets.

  We were using the nearest site neighbor distances to identify the 35 most remotely located / isolated mining sites as test sites for the mining site detection. The remaining 104 mining sites are used as training sites. Since some mining sites are very closely located, resulting in overlapping samples, we chose this selection strategy to minimize similarity effects and avoid overlapping samples in the train and test set.

  \begin{figure}[htbp]
    \centering
    \subfloat[Artisanal Mining Sites]{\label{fig:a}\includegraphics[width=\imgWidth]{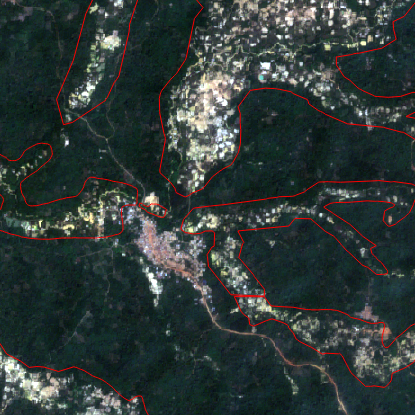}}\hfill
    \subfloat[Open Pit]{\label{fig:b}\includegraphics[width=\imgWidth]{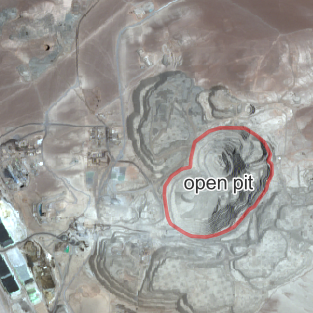}}\\[5mm]
    \subfloat[Mine Facilities]{\label{fig:c}\includegraphics[width=\imgWidth]{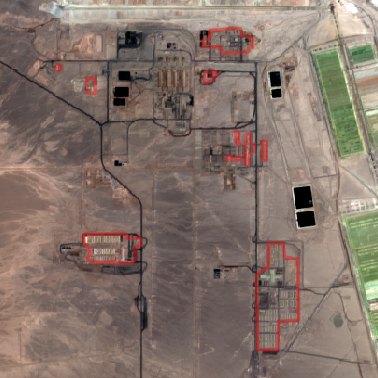}}\hfill
    \subfloat[Waste Rock Dump]{\label{fig:d}\includegraphics[width=\imgWidth]{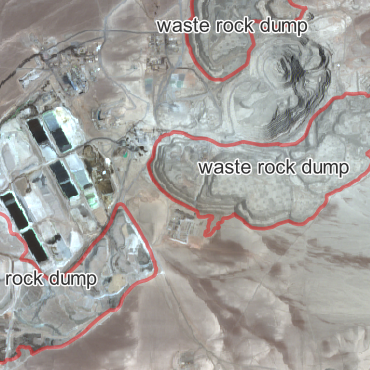}}\\[5mm]
    \subfloat[Stockyard]{\label{fig:e}\includegraphics[width=\imgWidth]{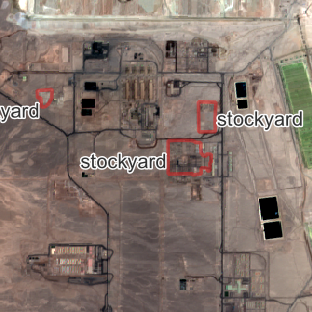}}\hfill
    \subfloat[Processing Plant]{\label{fig:f}\includegraphics[width=\imgWidth]{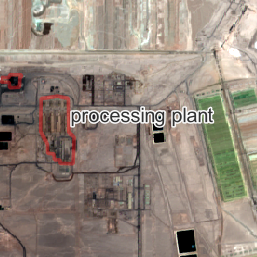}}\\[5mm]
    \subfloat[Tailings Storage Facility]{\label{fig:g}\includegraphics[width=\imgWidth]{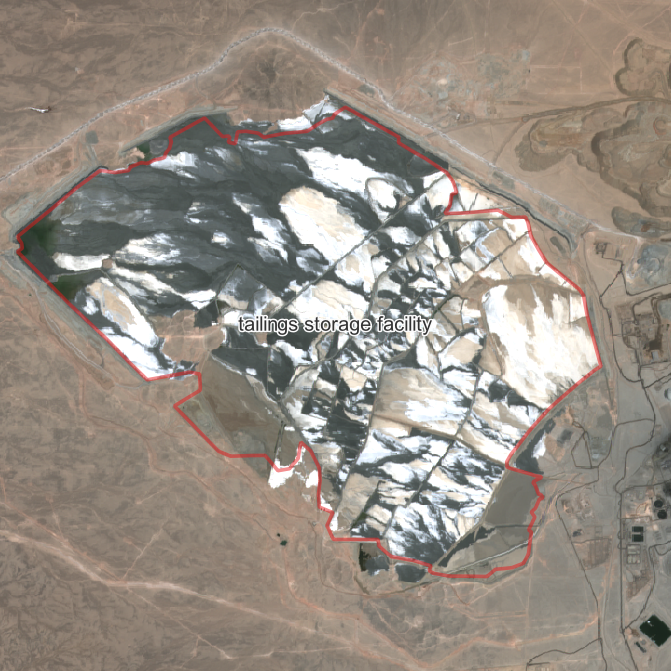}}\hfill
    \subfloat[Heap Leaching]{\label{fig:h}\includegraphics[width=\imgWidth]{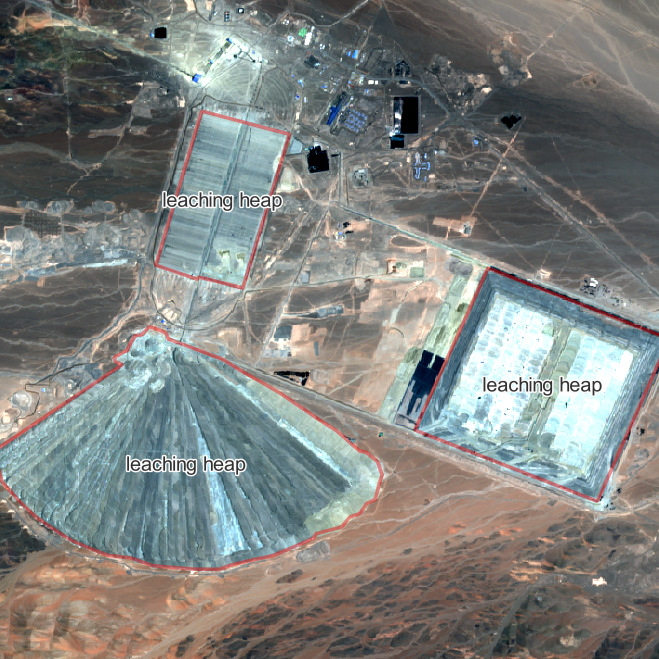}}
    \caption{Examples of the true color (RGB bands) Sentinel 2 crops of the considered Mining Site Sections.}
    \label{fig:sections_examples}
  \end{figure}

\subsubsection{Eminent Mining Site Sections} 

The following eminent sectors are of relevance in the surface mining context. These classes are represented with polygons in our data-set. Figure~\ref{fig:sections_examples} gives an impression of the classes appearance. 

\begin{description}[style=nextline, labelwidth=1mm, labelindent=0cm, leftmargin=!, itemsep=1em]
    \item[\textbf{Artisanal Mining (ASM)}] sites are distinguished as hand-crafted mining with simple tools. This kind of mining is usually associated with gold mining in Africa. 
    
    \item[\textbf{Large Scale Mining (LSM)}] sites are commercially run by companies, have an infrastructure, a processing chain, and heavy machinery used in the mining process.
    
    \item[\textbf{Open Pit}] defines the area where the material is moved from the ground. These pits are not necessarily active anymore and are sometimes used as waste rock dumps at the end of their life cycle.
    
    \item[\textbf{Mine Facilities}] are buildings or structures that are mainly for human comfort or decoupled from the mining process, including parking areas, shops, canteens, and laboratories.
    
    \item[\textbf{Waste Rock Dump}] is created to store the ore remaining after the extraction of all desired materials. These areas can grow to large hills and are usually close to the pit.
    
    \item[\textbf{Tailings Storage Facilities}] are used to collect and store the wastewater of the mine. This wastewater is charged with unusable concentrations of several chemicals and reagents that remain from the mining process. This wastewater is called tailings and needs to be kept in lake-like water bodies to prevent it from leaking into natural water bodies.
    
    \item[\textbf{Stockyards}] Large and remote mines often manage stockyards for redundancy of important equipment. Transportation of heavy equipment to remote places can be costly and time-consuming.
    
    \item[\textbf{Processing Plant}] Inside the processing plant, the desired materials are being extracted from the ore, concentrated, and isolated from any impurities, up to a certain degree. This process consists of multiple sub-processes that may distribute to individual buildings.
    
    \item[\textbf{Heap Leaching}] describes a chemical-based process to separate the desired material (e.g. metals, uranium) from the crushed ore. This process creates acid and other contaminated substances, and, in the case of uranium, also radioactive fluids~\cite{HardRockMiningHandBook}.
\end{description}

\begin{table}[htbp]
    \centering
    \footnotesize
    \caption{The table shows the selected mining site sections, the used key inside the data set, the number of occurrences in the considered mining sites, and their pixel value in the mask files.}
    \setlength{\tabcolsep}{2.5mm}
    \renewcommand{\arraystretch}{1.2}
    %\begin{tabular}{@{}lrrc@{}}
    \begin{tabular}{lrcc}
      \toprule
      Class                     & segments      & Key          & px mask            \\ \midrule
      other                     & -    & -            & \texttt{0}         \\
      ASM site                  & -    & \texttt{asm} & \texttt{1}         \\
      LSM site                  & 150  & \texttt{lsm} & \texttt{2}         \\
      open pit                  & 143  & \texttt{op}  & \texttt{5}         \\
      mine facility             & 416  & \texttt{mf}  & \texttt{4}         \\
      waste rock dump           & 320  & \texttt{wr}  & \texttt{9}         \\
      stockyard                 & 158  & \texttt{sy}  & \texttt{7}         \\
      processing plant          & 80   & \texttt{pp}  & \texttt{6}         \\
      tailings storage facility & 59   & \texttt{tsf} & \texttt{8}         \\
      heap leaching             & 31   & \texttt{lh}  & \texttt{3}         \\ \bottomrule
    \end{tabular}
    \label{tab:classes}
  \end{table}

  \def\imgHeight{45mm}
  \def\imgWidth{0.75\linewidth}

  % Figure 1
  \begin{figure}[t]
    \centering
    \includegraphics[width=\linewidth]{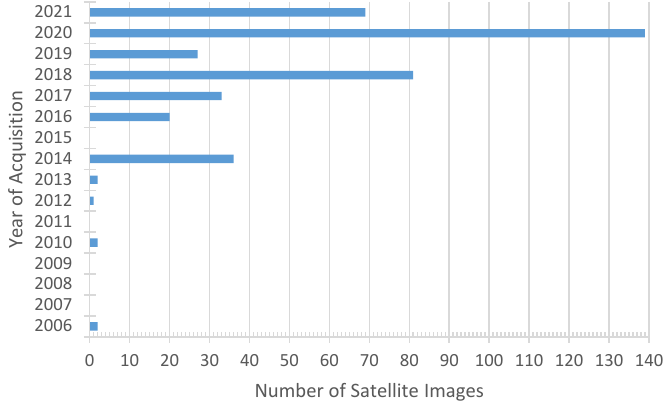}%
    \caption{Acquisition year of the hi-res satellite imagery that is used in the annotation process.}
    \label{fig:imagesperyear}
  \end{figure}

\subsubsection{Mining Facility Metadata}

  % Figure 2
  \begin{figure}[t]
    \centering
    \includegraphics[width=\imgWidth]{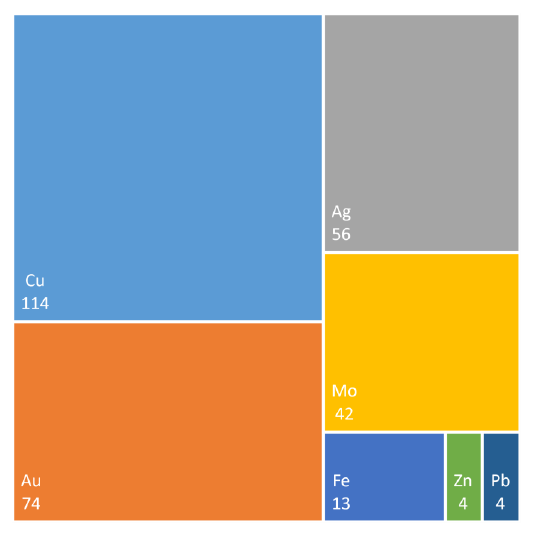}%
    \caption{The number of sites that extract the corresponding commodity.}
    \label{fig:numberofmines}
  \end{figure}

  A large mining site can have a life cycle over decades. The \texttt{status} describes the current state in this life cycle with different attributes like \textit{in operation}, already \textit{decommissioned}, \textit{in maintenance}, etc. The status of a mine is of high interest. It helps in understanding previous and future activities. The different \texttt{type}s of mining sites help understand the kind of environmental change and narrow the set of potentially used chemicals. While \textit{OpenPit} mining has a substantially different footprint and environmental impact as \textit{Underground mining} in comparison. The \texttt{processing method} narrows the amount of potentially consumed energy, water, used chemicals, and produced waste even further. The \texttt{ROM} (run of mine) regards the annual extracted ore in Megatons per year, where \texttt{production} represents the amount of the extracted and finally processed material (\texttt{commodity}) in tons per year. The \texttt{ore grade} shows the average percentage of the desired commodity in the ore. Further information on the involved companies and information sources is described in \texttt{shareholder} and \texttt{sources}.

  The mining sections have been annotated with the commercial product Google Earth in the resulting KML format. Google Earth represents satellite and aerial images of multiple commercial products. As a consequence, the \texttt{imaging system} and \texttt{imaging date} need to be saved to maintain replicability. Fig.~\ref{fig:imagesperyear} shows that most satellite images are recorded in the year 2014 and later, while 2020 is the year with the most images. Most annotations (1416) are based on satellite images that are accessible via Google Earth from \texttt{Maxar Technologies}, while 632 annotations are based on \texttt{CNES / Airbus} and the remaining 157 annotations are from both agencies.

% The main image source is with 1416 cases \texttt{Maxar Technologies}, in 623 cases \texttt{CNES / Airbus}, and in 157 cases a combined image of both.

  % Figure 3
  \begin{figure}[t]
    \centering
    \includegraphics[width=\imgWidth]{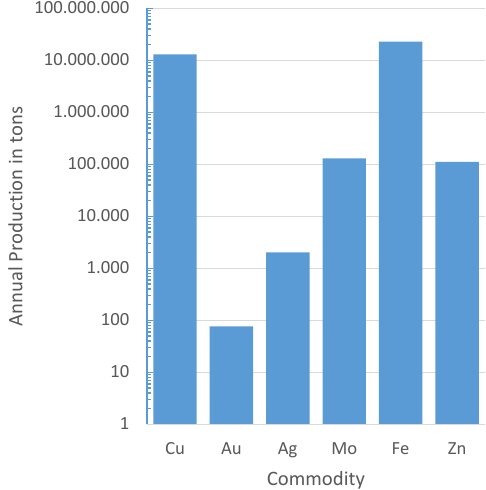}%
    \caption{Production in tons per year for the chosen LSM sites.}
    \label{fig:tonsperyear}
  \end{figure}

\subsection{ASM}\label{dsASM}

  While the locations of LSM sites are usually well known, ASM sites are growing unregulated and without awareness of authorities and therefore need to be found in drone, aerial photography, or satellite images. The main effort in the annotation process of ASM sites lies in finding recently created spots in the corresponding ASM hot spot areas: Western, Western North, Central, Ashanti, and Eastern Regions of Ghana. The selected search area comprises 9,711\,km² and lies mainly in the Ashanti and Western Region of Ghana, see Fig.~\ref{fig:asmsearcharea}. The ASM sites do not have clear boundaries, but rather form like branches or roots along rivers, creeks, and rills, see Fig.~\ref{fig:asmform}.

%%%%%%%%%%%%%%%%%%%%%%%%%%%%%%%%%%%%%%%%%%%%%%%%%%%%%%%%%%%%%
\section{Initial Experiments}
  
\begin{figure}[t]
    \centering
    \includegraphics[width=\linewidth]{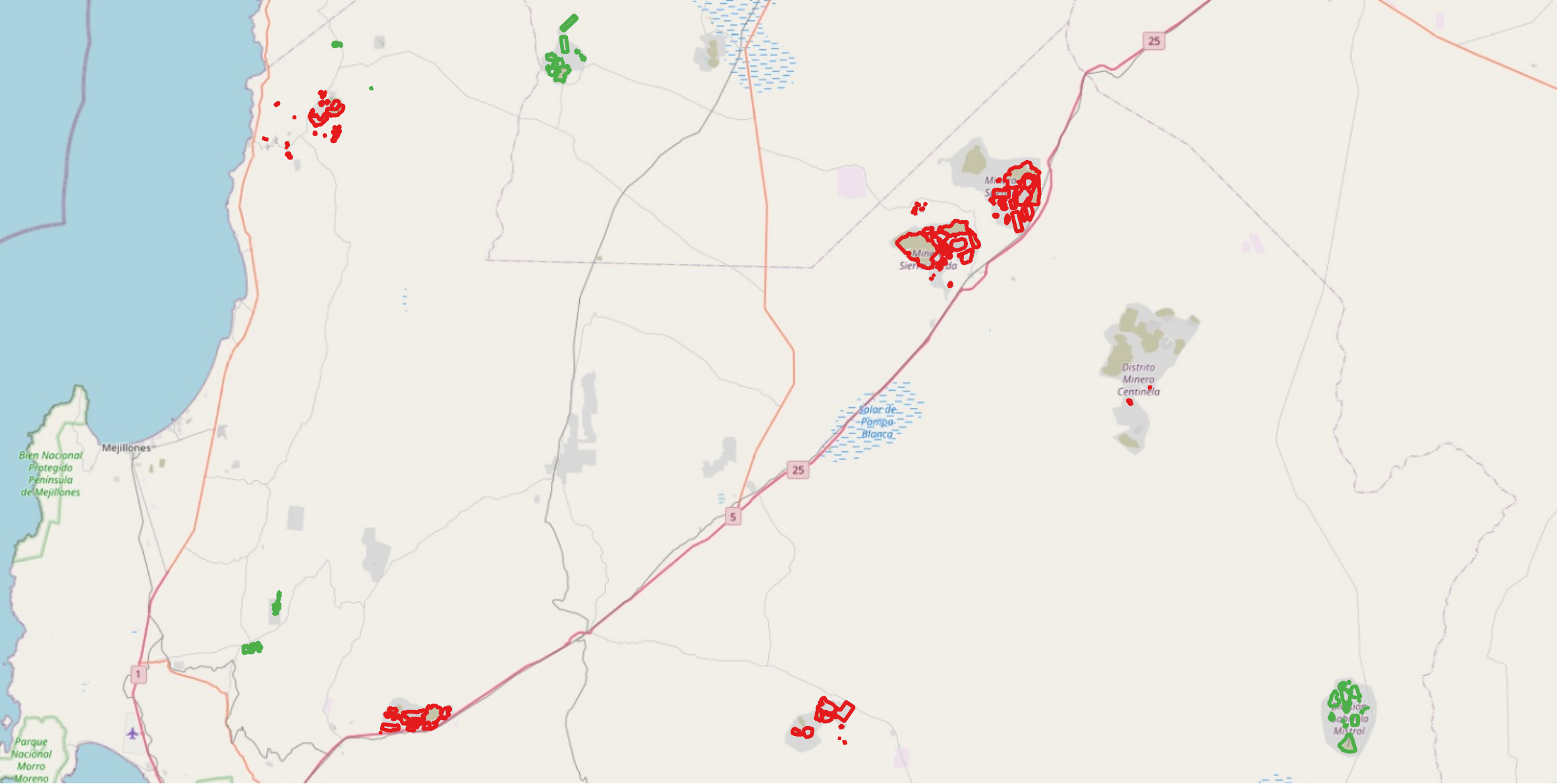}
    \caption{The map shows the central Antofagasta región, covering the mining sector of several training (red) and testing (green) sites.}
    \label{fig:train_test_map}
  \end{figure}

  Our initial experiments on the data set depict 2 different tasks on different imagery. The first task is a semantic segmentation on the mining site bounding box in hi-res satellite imagery (Maxar and CNES provided by Google Maps) to segment individual mining site sectors. The second task focuses on the detection of the mining site itself on a large scale (Country of Chile), based on mid-res satellite imagery (Sentinel-2). In both experiments, the training, validation, and test data sets are split based on entire mining sites rather than individual patches, ensuring that all patches from a given site are confined to either the training or test set, see Fig~\ref{fig:train_test_map}.

\subsection{Mining Sector Classification (HiRes Imagery)}

\begin{figure}[htbp]
  \centering
  \includegraphics[width=\linewidth]{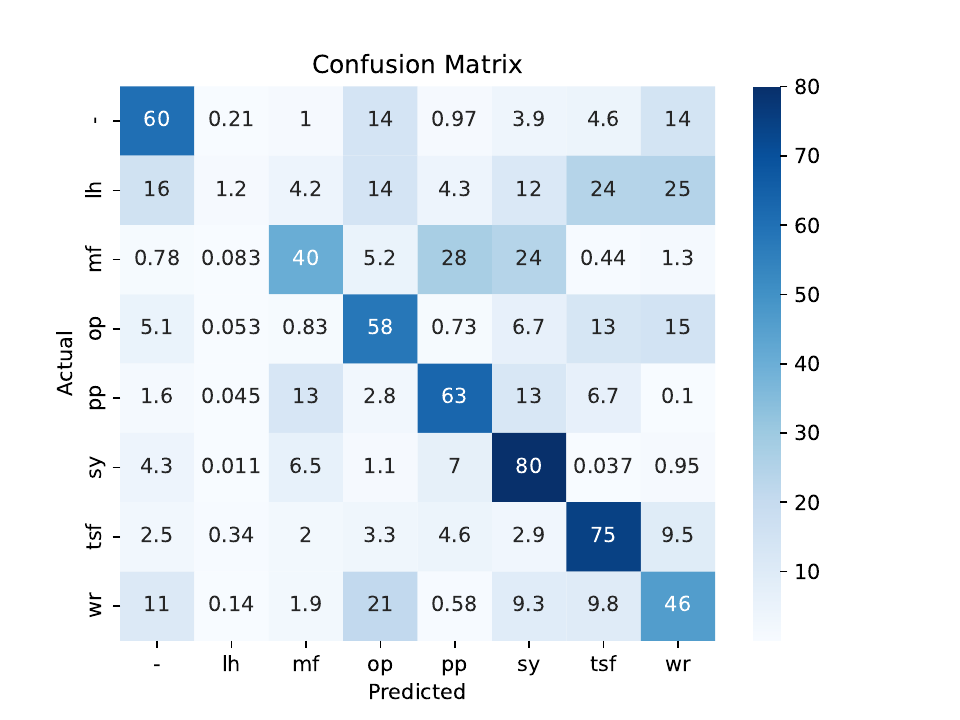}
  \caption{The confusion matrix shows that for most classes the detection rate lies above 50\,\%. Surprisingly, leaching heaps are more than 98\,\% misclassified as any other class.}
  \label{fig:cm_sector}
\end{figure}

  We selected the established U-Net architecture \cite{10.1007/978-3-319-24574-4_28}, incorporating a ResNet-50 backbone \cite{7780459} trained on ImageNet \cite{DenDon09Imagenet} as the network architecture. U-Net is a widely recognized semantic segmentation model, demonstrating robust performance in both computer vision and remote sensing applications. The mining sites were divided into 38 for training, 14 for validation, and 19 for testing. Each bounding box of the mining sites was divided into patches measuring 256\,$\times$\,256 pixels. We applied standardization for normalization of the input images. Model training was conducted using the cross-entropy loss function with the Adam optimizer. We employed a cyclical learning rate strategy with a \texttt{triangular2} mode scheduler. In this configuration, the scheduler linearly varies the learning rate from a base of 10\textsuperscript{-4} to a maximum of 10\textsuperscript{-3} over 2000 iterations, and then back down to the base rate over the same number of iterations, forming a complete cycle every 4000 iterations. After each cycle, the scheduler halves the maximum learning rate, resulting in progressively narrower oscillations. This approach helps the model explore a wide range of learning rates early on—potentially escaping suboptimal local minima—and then gradually refines the search space, leading to more stable convergence as training proceeds. To address substantial class imbalance, we applied class-specific pixel weighting derived from training set statistics.

\begin{figure}[htbp]
  \centering
  \includegraphics[width=0.85\linewidth]{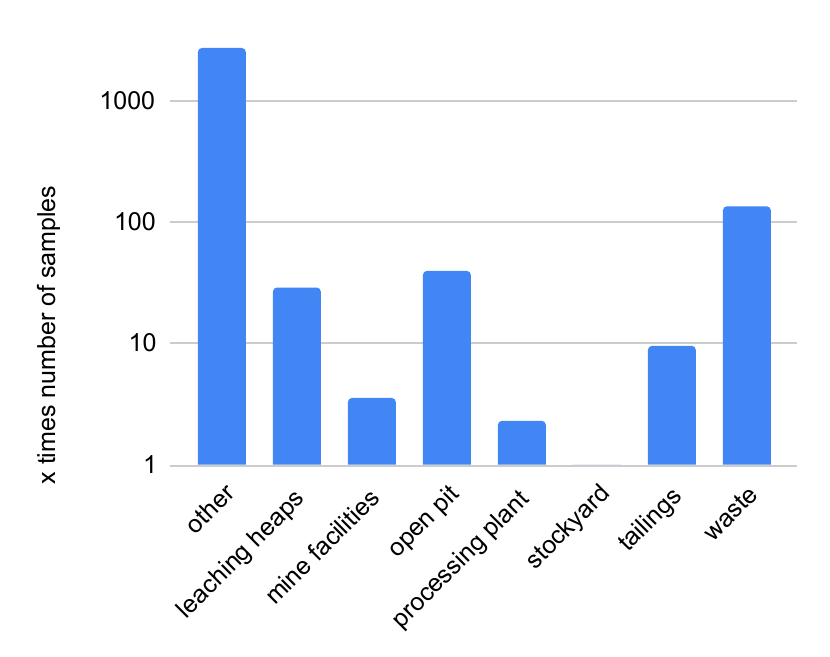}
  \caption{The class stockyard has the least number of samples. The figure represents the tremendous class imbalance, that makes class weighting during training an essential step.}
  \label{fig:samples_sector}
\end{figure}

\iffalse
  \XZ{here the readers would expect much more discussion about the results, e.g.. 
  \begin{itemize} \setlength\itemsep{0mm}
    \item average accuracy, overall accuracy
    \item which classes are particularly difficult to be correctly classified and why
    \item Whether the achieved accuracy can serve real world applications, if yes why; if not, what else is necessary. Which methods could potentially address these
  \end{itemize}}
\fi

\iffalse
\begin{table}[htbp]
\caption{Classification performance using the ImageNet pre-trained backbone.\label{tab:performance}}
\centering
\setlength{\tabcolsep}{15pt}
%\renewcommand{\arraystretch}{1.1}
\begin{tabular} {lcccc}
    \tophline
    \textbf{Class} & \textbf{Accuracy} & \textbf{Precision} & \textbf{Recall} & \textbf{F1-Score}\\
    \middlehline
    \texttt{0}: Unlabeled& 92.36& 0.90& 0.92& 0.91\\
    \texttt{2}: LSM&   89.50& 0.92& 0.89& 0.91\\
    \texttt{3}: Leaching Heap& 97.27& 0.93&  0.97& 0.95\\
    \texttt{4}: Mining Facilities& 61.79& 0.59&  0.62& 0.61\\
    \texttt{5}: Open-pit&   92.40& 0.94& 0.92& 0.93\\
    \texttt{6}: Processing Plant& 63.06& 0.89& 0.63& 0.74\\
    \texttt{7}: Stockyard& 32.43& 0.65& 0.32& 0.43\\
    \texttt{8}: Tailings Storage Facility& 96.26& 0.87& 0.96& 0.92\\
    \texttt{9}: Waste Rock Dump& 92.44& 0.87& 0.92& 0.90\\
    \bottomhline
\end{tabular}
\label{tab:performance}
\end{table}
\fi

\subsection{Mining Site Detection (S2 Imagery)}
  This experiment is intended to reflect the suitability of the data set for mining site detection. In this 2-class case, all mining site sectors are combined to the class \emph{mine-site}. The opposite class includes all remaining surfaces. Due to the ground sampling distance of 10\,m of the Sentinel-2 imagery, the available training surface is with 297 training, 131 test, and 61 validation patches small, leading to rather unsatisfactory results with a lot of miss-classifications in favor of the \emph{no-mine} class. To increase the heterogeneity of the \emph{no-mine} class in the training data, we randomly selected 10,000 sample patches from the whole Chile set with the assumption of being \emph{no-mine}. The reason for that is the very low probability of hitting a mining site, since around 0.5\,\% of Chile's area is associated with mining sites, based on the area of Chile (756,101\,km²) and the sum area of all Chilean mining sites (3,760\,km²) according to \cite{maus2020gmpv}. By this blind assumption, we could reduce our false negative rate from 36\,\% to 25\,\%, which led to a similar increase of FP rate from 6.1\,\% to 9.4\,\%.

\subsection{Results}

  We show common classification performance metrics micro and macro averaged. Especially in the mining sector classification, the extreme class imbalance challenges a proper evaluation and demands differently aggregated metrics, according to the specific use case. All metrics are computed at pixel level.

  \textbf{Mining Sector Classification:}
    The mean micro accuracy over all samples on the test set is 59.2\,\%. Across classes, the macro accuracy is 89.0\,\%. As shown in Fig. \ref{fig:cm_sector}(a), most classes are classified with over 50\,\% accuracy; however, the network struggles with the \texttt{lh} class, representing leaching heaps, with 98\,\% misclassified as other classes, leading to the assumption that leaching heaps might look very differently across mining sites. Surprisingly, \texttt{sy}, representing stockyard, is classified decently despite its low representation in the dataset, leading to the assumption that stockyards are less heterogeneous represented across different mining sites. Simply saying, that stockyards are similar for each mining site. 

  It also highlights the importance of our class weighting strategy. 
  This baseline model already demonstrates promising results and potential for real-world applications with the introduced dataset. 

  \textbf{Mining Site Detection}
    The outcome of this experiment shows a fair result with a very similar performance by using the 10,000 random selected Chile patches as \emph{no-mine}. Therefor in this case, the augmentation with pseudo-labeling of unlabeled data has no significant advantage. Since this is a binary classification, an accuracy of 0.89\,\% and an F-Score of 0.75 are rather poor and only a little if no help in a real world scenario. 

  \begin{table}[htbp]
    \footnotesize
    \centering
    \setlength{\tabcolsep}{1.35mm}
    \renewcommand{\arraystretch}{1.2}
    \begin{tabular}{r|cccc|cccc}
                        & \multicolumn{4}{c}{Macro Average}      & \multicolumn{4}{c}{Micro Average}      \\
    Task                & Acc. & Prec. & Rec. & F1     & Acc. & Prec. & Rec. & F1     \\ \midrule
    Sector Class.       & 0.89    & 0.17     & 0.53  & 0.17  & 0.59    & 0.59     & 0.59  & 0.59  \\
    Site Detect.        & 0.91    & 0.74     & 0.79  & 0.76  & 0.91    & 0.91     & 0.91  & 0.91  \\
     +10k               & 0.89    & 0.71     & 0.83  & 0.75  & 0.89    & 0.89     & 0.89  & 0.89 
    \end{tabular}
  \end{table}
    
  \textbf{Results in Context}
    We still see the yet non-satisfactory results as a proof that a detection of mining sites and the classification of eminent mining site sectors on very homogeneous, mostly non-vegetated grounds, as present in Chile, is possible and can likely significantly improved with state-of-the-art fine tuning, model training, augmentation etc.

  \begin{figure}[htbp]
    \centering    
    \subfloat[Results of the Mining Site detection.]{\includegraphics[width=0.51\linewidth]{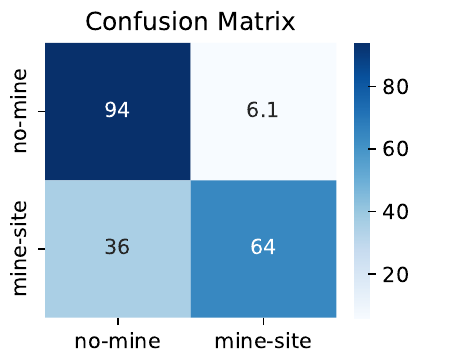}}%
    \hfill
    \subfloat[Results of the Mining Site detection, including 10,000 randomly selected Chile patches, labeled as \emph{no-mine}]{\includegraphics[width=0.48\linewidth]{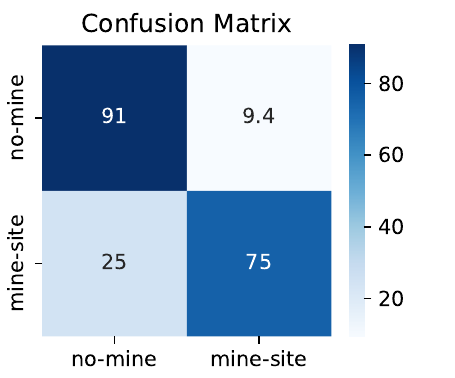}}%
    \caption{Macro Average (class-wise) Confusion Matrix in percentages of the Mining Site Detection experiment.}
  \end{figure}

%%%%%%%%%%%%%%%%%%%%%%%%%%%%%%%%%%%%%%%%%%%%%
\section{Applications and Enrichment}\label{useCases}

% As discussed in section 2.3 (impact on artisans' livelihood), there is an urgent need to 

\par
We envision the following potential use cases of our data set:
\begin{enumerate}
    \item Machine learning-based mineral mapping will be further promoted by the proposed data set. The data set covers multiple mined ores with appropriate labels, thus allowing the machine learning models to be trained for those models.
    
    \item Mining site monitoring will become further easier, with the detailed annotations made available by this data set. AI on Mining sites is not just about mineral mapping, it also involves a variety of infrastructures. Our data set, for the first time, makes publicly available different eminent mining site sectors. This will be implicitly useful in case of any disaster/accident management involving the mining site.
    
    \item Our data set will also act as a precursor for further data sets, including in the future.
    
    \item Our data set covers potentially environmentally harmful sectors like water bodies and leaching heaps. Such classes can be helpful to study vegetation stress and predict any future disaster involving water in the mining area.
    
    % \item Our data set can be studied in conjunction with available weather data to study the effect of mining on climate change. % that is way too broad
    
    \item To prevent violation of human rights, especially of artisanal miners who traditionally rely on such activities for their livelihoods, the data set can also support constructive remediation and restoration measures to reverse or minimize the negative externalities associated with ASM. Indeed, such restoration and remediation measures may be easier in the case of small-scale mining than in LSM \cite{haddaway2019evidence}. 
\end{enumerate}

%%%%%%%%%%%%%%%%%%%%%%%%%%%%%%%%%%%%%%%%
\section{Conclusions}\label{conclusions}

AI-driven remote sensing is a milestone in identifying and monitoring environmental degradation. The impact of mining facilities on the natural environment can be enormous. Monitoring mining activities and their environmental impact is a crucial step toward reducing the negative impact on the natural environment. Our data set consists of 150 LSM sites and their eminent sectors, including metadata and already processed multi-spectral satellite imagery. The data set may allow the training of AI-driven models that enable mineral detection, quantity approximations, estimation of the environmental impact, etc. In the future, we plan to further enrich our dataset with hyperspectral images.

%%%%%%%%%%%%%%%%%%%%%%%%%%%%%%%%%%%%%%%%%%%%%%
%\section{Acknowledgment}\label{acknowledgment}

%% The following commands are for the statements about the availability of data sets and/or software code corresponding to the manuscript.
%% It is strongly recommended to make use of these sections in case data sets and/or software code have been part of your research the article is based on.

%\codeavailability{TEXT} %% use this section when having only software code available

%\dataavailability{TEXT} %% use this section when having only data sets available

The Sentinel-2 based mining site Patches are available in a HuggingFace Repository, the experimental setup and the annotation polygons as well as auxiliary tools are available on GitHub:
\begin{itemize}
    \item LAMES Sentinel-2 mining site patches: https://doi.org/10.57967/hf/7223 \cite{lamesimages}
    \item LAMES Dataset (annotations, shapes, metadata, etc.): https://doi.org/10.57967/hf/7289 \cite{lamesmetadata}
    \item LAMES baseline experimental setup: https://github.com/zhu-xlab/mineseg
\end{itemize}
 %% use this section when having data sets and software code available

Some samples are available here: https://github.com/chenzhaiyu/mineseg/sample\_data %% use this section when having geoscientific samples available

%\videosupplement{TEXT} %% use this section when having video supplements available

\appendix

\section{Folder structure}    %% Appendix A

% Make it compact + monospaced
\renewcommand*\DTstyle{\ttfamily\footnotesize}
\setlength{\DTbaselineskip}{10pt} % e.g., tighter lines (or use an absolute length like 10pt)
\DTsetlength{10pt}{5pt}{5pt}{1pt}{3pt}
\begin{figure}
    \flushleft
    \dirtree{%
        .1 project/.                                            
        .2 annotations/.                                        
        .3 annotation\_doc.pdf.                                 
        .3 Chile\_LSM\_sectors.geojson.                         
        .3 Chile\_LSM\_sites\_Maus\_et\_al\_subset.geojson.     
        .3 Ghana\_ASM.geojson.
        .3 overview.qgs~.
        .3 overview.qgz.
        .3 test\_sites.geojson.
        .3 train\_sites.geojson.
        .2 example Images/.
        .3 hires\_msk\_patch.png.
        .3 hires\_tci\_patch.png.
        .3 lh/.
        .3 s2\_msk\_patch.png.
        .3 s2\_msk\_patch.tif.
        .3 s2\_tci\_patch.png.
        .3 s2\_tci\_patch.tif.
        .3 train\_test\_sites\_map\_lowres.pdf.
        .3 train\_test\_sites\_map.pdf.
        .2 img\_sector/.
        .3 multiclass\_image\_data/.
        .3 multiclass\_image\_data.zip.
        .2 img\_site/.
        .3 image-data.tar.gz.
        .2 metadata/.
        .3 metadata.csv.
        .3 metadata\_sources.txt.
        .3 metadata.xlsx.
        .2 results/.
        .3 sector\_classification.txt.
        .3 site\_detection\_10k\_rand.txt.
        .3 site\_detection.txt.
        }
    \caption{Caption}
    \label{fig:placeholder}
\end{figure}

%\subsection{}     %% Appendix A1, A2, etc.

%\noappendix       %% use this to mark the end of the appendix section. Otherwise the figures might be numbered incorrectly (e.g. 10 instead of 1).

%% Regarding figures and tables in appendices, the following two options are possible depending on your general handling of figures and tables in the manuscript environment:

%% Option 1: If you sorted all figures and tables into the sections of the text, please also sort the appendix figures and appendix tables into the respective appendix sections.
%% They will be correctly named automatically.

%% Option 2: If you put all figures after the reference list, please insert appendix tables and figures after the normal tables and figures.
%% To rename them correctly to A1, A2, etc., please add the following commands in front of them:

%\appendixfigures  %% needs to be added in front of appendix figures

%\appendixtables   %% needs to be added in front of appendix tables

%% Please add \clearpage between each table and/or figure. Further guidelines on figures and tables can be found below.

\paragraph{Author Contribution} Conceptualization and investigations were carried out by MaKa, SuSa (technically), and MrKo (law, social science and ethical perspectives). ZhCh and MaKa designed, implemented, and evaluated the experimental setup. MaKa contributed visualizations. XXZ provided the financial opportunity. MaKa, SuSa, MrKo, ZhCh, LuKo, and XXZ contributed several textual passages. The geospatial annotations and metadata were contributed by the Institute of Mineral Resources Engineering (MRE) of RWTH Aachen. Language and content review and editing has been applied by MaKa, SuSa, MrKo, ZhCh, XXZ. Funding acquisition: LuKo, XXZ %% this section is mandatory

%conceptualizing, investigating, writing and reviewing were from a law, social science and ethics perspective

\paragraph{Author Contribution}MaKa, SuSa, and MrKo conceptualized the papers content. ZhCh and MaKa conceptualized and implemented the experimental setup and ran evaluation. MaKa contributed visualizations. Scientific Investigations by MaKa, SuSa, and MrKo. XXZ provided the financial opportunity. MaKa, SuSa, MrKo, ZhCh, LuKo, and XXZ wrote and contributed textual passages. Conceptualizing, investigating, writing and reviewing were from a law, social science and ethics perspective The annotations and metadata were contributed by the Institute of Mineral Resources Engineering (MRE) of RWTH Aachen. Review and editing: MaKa, ZhCh, SuSa, MrKo, XXZ. Funding acquisition: LuKo, XXZ %% this section is mandatory

\paragraph{Competing Interests} The contact author has declared that none of the authors has any competing interests. %% this section is mandatory even if you declare that no competing interests are present

%\disclaimer{TEXT} %% optional section

\paragraph{acknowledgements}
The annotation process was conducted by the Institute of Mineral Resources Engineering (MRE). This work was supported by DynamicEarthNet and the Future Lab AI4EO.

%% REFERENCES

%% The reference list is compiled as follows:

%\begin{thebibliography}{}
%\bibitem[AUTHOR(YEAR)]{LABEL1}
%\end{thebibliography}

%% Since the Copernicus LaTeX package includes the BibTeX style file copernicus.bst,
%% authors experienced with BibTeX only have to include the following two lines:
%%
\bibliographystyle{copernicus}
\bibliography{references}

\end{document}